# Problematizing AI Omnipresence in Landscape Architecture

Phillip Fernberg[1,2], Zihao Zhang[3]

[1]Utah State University, Utah/USA · phillip.fernberg@usu.edu
[2]OJB Landscape Architecture, California/USA
[3]City College of New York, New York/USA

**Abstract:** This position paper argues for, and offers, a critical lens through which to examine the current AI frenzy in the landscape architecture profession. In it, the authors propose five archetypes or mental modes that landscape architects might inhabit when thinking about AI. Rather than limiting judgments of AI use to a single axis of acceleration, these archetypes and corresponding narratives exist along a relational spectrum and are permeable, allowing LAs to take on and switch between them according to context. We model these relationships between the archetypes and their contributions to AI advancement using a causal loop diagram (CLD), and with those interactions argue that more nuanced ways of approaching AI might also open new modes of practice in the new digital economy.

## 1 Introduction

Amidst the ever-evolving discourse on technology and design, the role of Artificial Intelligence (AI) in shaping the contours of landscape architecture (LA) processes has become a significant industry trend (FERNBERG & CHAMBERLAIN 2023). This intersection, rich with both potential and pitfalls, necessitates a nuanced understanding of how human designers think about AI's impact and the diverse ways it interweaves with the state of the practice. While tech-focused researchers and practitioners have been investigating subjects under the AI umbrella for some time, it has only recently hit the mainstream consciousness; largely due to the explosion of generative tools such as ChatGPT, Midjourney, and Stable Diffusion. After a few years of debate, experimentation, and contemplation, society's enthusiasm for generative tools has shifted from initial excitement to introspection and reassessment. Jackie Fenn and colleagues might suggest the generative hype cycle is reaching its so-called *trough of disillusionment* (FENN & BLOSCH 2018). AI was the 2023 buzzword of the year (BHATTACHARYA 2023) and captured an inevitable ubiquity in which it seems to be everywhere and nowhere all at once. Most seem to know the term but many are only familiar enough to default to popular culture caricatures of singular intelligent agents operating at odds with human will and creativity (CANTRELL & ZHANG 2018). Moreover, we posit the current AI discourse is at once too one-dimensional and too focused on judging the merits of AI-driven technology rather than critiquing the way humans think about and use it. To overcome this myopia requires diversify perceptions of AI and our relationship to it.

Discourse on human-tech relations are well established in architecture and aesthetics, including work like Negroponte's speculation on emergent architectural machines in computer-aided design or CAD (NEGROPONTE 1970) or technology philosopher Don Ihde's models of the personal computing revolution of the 1990s (IHDE 1990). Building on previous theories, our work addresses the specific problems of human-AI relations in the context of landscape architectural practice. In this position paper, we introduce various AI-LA archetypes representing different perceptual lenses landscape architects (LAs) might take toward

AI development; describe the benefits and drawbacks each one brings to praxis; and then model interactions between these archetypes. In in doing so, we argue how their interactions can inspire how the industry might adapt, resist, or reshape itself in response to AI's influence. These polymodal archetypes are not mutually exclusive identities but rather permeable mental modes one can find themselves moving between at any given time.

## 2 Problematization: AI-LA Archetypes and Narratives

### 2.1 The Optimizer

The quintessential "move fast and break things" persona, The Optimizer believes that AI represents an aggregate of tools evolved over time to help LAs get better, smarter, and more efficient, and should be accelerated at all costs. To them AI is the bridge to bypass busy work (think redrafting, sheet formatting, and image post-production) for more billable hours spent on the meaningful parts of the design process. The most valuable byproduct for optimizers is being able to do *more* of their most valuable work *faster*. Such could mean happier employees, better products, and outpacing the competition. Examples of Optimizers in action are the companies using websites like Civitai.com where enthusiasts of Stable Diffusion share their checkpoint and Lora models. These smaller, fine-tuned models work like libraries that can be added to basic Stable Diffusion to perform special tasks. For example, one can fine-tune their Stable Diffusion and create a Lora model that mimics one artist's art style. At the multinational firm SWA, a team of designers has trained a Lora model that can produce hand-sketch drawings in the style of one of their Principals (DOMLESKY et al. 2023). Quickly, we will expect more firms with the resources to adopt their own proprietary AI systems to optimize their own workflow.

When developing AI-LA tools, Optimizers must be sure to examine the reality of technological determinism that underpins their effort to "revolutionize" the LA industry. They are often prone to rosy views of technological advancement, and dismissive of its accompanying challenges. An Optimizer might assume for instance that with AI people will inevitably work fewer hours per week – the opposite is often true –- and overlook the busy work or bureaucracy created by new AI workflows – humans still have to learn how to interact with AI systems and manage their workloads. Furthermore, the development of AI systems, often framed as inclusive and enabling, can create other inequities among firms, institutions, and individuals. Take the example of accessibility, where only those individuals and organizations with extra resources and capacity can afford subscriptions to "test out" the newest and most advanced generative tools. Software subscriptions can quickly add up in firm budgets or get lost in University bureaucracies, as was recently the case when Author Zhang attempted to use studio funds for giving students access to Midjourney. Optimizers should realize that AI should not only optimize efficiency and speed, but other factors, like workers' wellbeing and equity.

### 2.2 The Resistor

The Resistor mindset sees the current wave of AI tech as a threat to creative agency and human thriving. Overreliance on AI dilutes humanist endeavors in the eyes of the Resistor. The art piece "Théâtre D'opéra Spatial" embodies the sort of provocation that puts Resistors to work. Created using Midjourney, Jason Allen submitted a to the Colorado State Fair's annual art competition using name "Jason M. Allen via Midjourney" and won the blue ribbon for emerging digital artists. While Allen and other artists like him have been unabashed in their machine-assisted methods, their achievements were offset by backlash and mass protest of other artists against AI-generated artwork in late 2022. Perhaps, a milestone was the case of "Zarya of the Dawn," a comic book generated using Midjourney that was first granted copyright in late 2022, and then revolved in February 2023 after the US copyright office found out the images were AI-generated. The following year, the US Copyright Office held a series of public hearings about rules and policies around generative AI content's copyright. In late October, a federal judge in California dismissed a lawsuit by visual artists who accuse Stability AI, Midjourney, and DeviantArt of misusing their copyrighted work in connection with the companies' generative AI systems (See Table 1 for other cases and hearings regarding copyright and intellectual property of AI generated content).

**Table 1:** Sampling of AI-related court cases and hearings through 2023

| Case/Hearing | Date | Summary |
| --- | --- | --- |
| Thaler v. Perlmutter | August 2023 | U.S. District Court ruled against copyright protection for AI-generated art, upholding human authorship requirement for copyright. |
| OpenAI Copyright Case | July 2023 | Legal challenge to OpenAI regarding the use of copyrighted materials for training AI models, with fair use as a key argument. |
| Getty Images v. Stability AI | January 2023 | Getty Images filed a copyright claim against Stability AI in the U.K. High Court over the use of copyrighted images in AI training. |
| U.S. Copyright Office Initiative | March 2023 | Initiative launched to examine AI's impact on copyright law and policy, including scope of copyright in AI-generated works and use of copyrighted materials in AI training. |
| Zarya of the Dawn | February 2023 | The US Copyright office believed that images in the work that were generated by the Midjourney technology are not the product of human authorship, thus canceling the original copyright certificate. |
| House Judiciary Subcommittee Hearing on AI and Copyright Law | May 2023 | Examined the intersection of AI and copyright law, focusing on the use of copyrighted works in AI training and copyright protection for AI-assisted works. |
| Senate Committee Hearing on AI and Copyright | July 2023 | Discussed AI's role in copyright matters, implications for various industries, and issues like fair use and metadata solutions in AI training models. |
| Artists' AI Copyright Lawsuit v Midjourney, Stability AI | October 2023 | A U.S. judge pared down a lawsuit filed by artists against Midjourney and Stability AI, related to the use of copyrighted materials in AI-generated art. |
| Beijing Internet Court Recognizes Copyright in AI-Generated Images | November 2023 | The Beijing Internet Court issued a decision recognizing copyright in AI-generated images. |

Many landscape designers also worry as the plaintiffs of intellectual property cases that the images of their designed landscape will be used in training generative AI models. They worry that AI makes design cheap and generic, perpetuates bias, or possibly infringes copyright (BELESKY 2023). The underlying logic of the tension between these Resistors and generative AIs is likely the crisis of authorship. Uncarefully trained generative AI models overtly erase what Foucault would call the "author function", leaving creative work without a name to whom to attribute meaning and liability, and perhaps most importantly, to give social credit to for its accomplishment (FOUCAULT 1969).

However, as it may be necessary to pin down an author for its function in legal purposes, we invite the Resistors to embrace an assemblage perspective, which might provide a sense of ease. Assemblage thinking posits that humans have always been co-evolved with other non-human actors around us, including tools and languages, and thus what we thought to be human agency and creativity has always been "distributive," to use Jane Bennett's term. In landscape architecture, AI tools are merely another type of actor that further diversify and hybridize the already "polluted" human creativity. Just as the profession has come to assert the integral role of ecological indeterminacy driving concepts – think of projects like SCAPE's Living Breakwaters and Field Operation's Fresh Kill Park where designers' agency and creativity is dissolved in a network of more-than-human actors – so too can Resistors accept a unique role for AI in the creative actor network.

## 2.3   The Stoic Instrumentalist

While colleagues buzz with either enthusiasm or apprehension about AI's role in their field, the Stoic approaches the change with a pragmatic indifference. They regard AI as just another tool to aid in their day-to-day tasks of drafting and designing landscapes. Even after living through the previous eras of 'Photoshopization' or Parametricism, the romance of things like manual sketching still holds value for them, yet they neither resist nor embrace the digital transition. AI is no different; each workday, they dutifully utilize their AI-assisted design software without fanfare, appreciating its utility but not swept up in the novelty. In a workspace where others are either thrilled or threatened by the encroaching technology, their apathetic stance towards the digital wave is a quiet divergence from the norm – or so they think as this is likely the most normative archetype of the bunch. For them, technology, encapsulated in the form of AI, is merely a facilitator, not a revolutionizer, in their enduring pursuit of shaping beautiful spaces and places. This sentiment has been evident in places like online forums, comment sections, or discussion spaces in public lectures, where a common reaction to AI provocations might be that it is "just another tool" in the toolbox (SALDANA OCHOA et al. 2023). To mitigate this indifference, the Stoics mentality needs reflection such as that of Science Technology and Society (STS), a field of study meant to open the black box of "ready-made" technology by showing the development of technical artefacts was historically contingent to social values and biases (BIJKER et al. 1987). STS has put more responsibility on society to regulate and influence the development of powerful technology like AI. By the same token, the Stoic mentality must shift to engage AI as a culture rather than merely a tool, and make sure our voices as landscape architects are heard in this AI revolution.

## 2.4   The Superuser

With grounding from Randy Deutsch's book *Superusers* on the emergence of technology specialists in the AEC industry (DEUTSCH 2019), this archetype navigates AI acceleration with a balanced blend of curiosity and practicality. The superuser approach to AI mirrors their approach to any new technology; they are tinkerers at heart, delving into the mechanics, potentials, and limitations of the tools at their disposal. Computational thinkers at their core, superusers distinguish themselves from other archetypes by co-evolving with technology rather than didactically leveraging (Optimizers), reacting to (Resistors), regulating (Protectors), or showing indifference to it (Stoics). In their hands, AI becomes a facilitator and partner, aiding in the efficient translation of creative ideas into tangible designs. Superusers might be embedded in firms, academic circles, or running their own technology-focused consultancies, but they are also convenors. They share their explorations and findings, both within their workplaces and external peer networks, fostering a collaborative culture of examining technology in the profession. The greater collective of ongoing JoDLA contributors is an emblematic group of Superusers (ERVIN 2022). While the ingenuity of the Superuser mental mode is generally a benefit to any technological reflection, it can also suffer from a myopia that favors tinkering with tools and methods over the more nuanced management of people and processes that are affected by use of those tools or methods. If the Superuser does not have counterbalancing, they may fail to see the bigger picture of how an AI intervention is to scale at the team, classroom, firm, or department level.

## 2.5 The Protector

The "Protector"'" archetype is akin to the caricatures of AI ethicists and safety advocates, representing a fusion of technical savvy with a deep commitment to safety, equity, and ethical considerations in the face of AI advancement. The Protector is a cautious integrator, meticulously balancing the innovative potential of AI with the imperatives of environmental stewardship and social justice. They may take positions like those of Timnit Gebru, Fei Fei Li, Nick Bostrom, or Kate Crawford, whose critiques emphasize the need for transparency in AI, the importance of diverse perspectives in technology development (BENDER et al. 2021) , and understanding AI as a sociotechnical system, where its impacts are as much social and environmental as they are technical. Designers in this archetype operate with the assumption of Helen Armstrong that AI has already transformed our profession and that it "is going to steamroll right over us unless we jump aboard and start pulling the levers and steering the train in a human, ethical, and intentional direction" (ARMSTRONG 2021). Their approach is not just about harnessing AI for efficiency or creativity in design but ensuring that these technological advancements serve broader societal goals, protect natural ecosystems, and promote inclusivity and fairness in both urban and natural landscapes. They are the guardians of ethical AI integration in landscape architecture, always vigilant about the potential consequences of AI on communities and the environment. This mental mode is evident in the work of educators such as Marc Miller at Penn State University or Charles Waldheim of Harvard who use AI to help students put a critical lens on the pastpresent, and future of the profession (MILLER 2022; WALDHEIM 2022). Protector mode was also at work in both LA students and practitioners throughout the past year in public venues such as conference sessions, webinars, and symposia with concerns on all of these issues well-articulated (DOMLESKY et al. 2023; SALDANA OCHOA et al. 2023).

## 3 Modeling Interactions between Archetypes

While we have cited work, firms, and individuals to help characterize the AI-LA archetypes, they should not be understood as fixed identities but rather mutable mental modes or perceptual lenses which can be taken on at any given time by any person, including those previously cited. Here we explain how those modes interact with each other and contribute to or impede AI advancement. An effective representational method is the causal loop diagram (CLD), developed and popularized in a subfield of systems theory called System Dynamics. System Dynamics is a modelling approach to understanding and simulating the behavior of complex systems over time. It has found use in anything from industrial processes to business and policymaking (MEADOWS 2008). A CLD is a system dynamics tool typically used to visually depict feedback loops and interdependencies among variables in a system. It consists of nodes representing the entities (in this case the archetypes) and arrows indicating causal relationships between them. Those arrows also include notations of polarity, which distinguish between reinforcing loops (positive or negative feedback) and balancing loops (both positive and negative feedback).

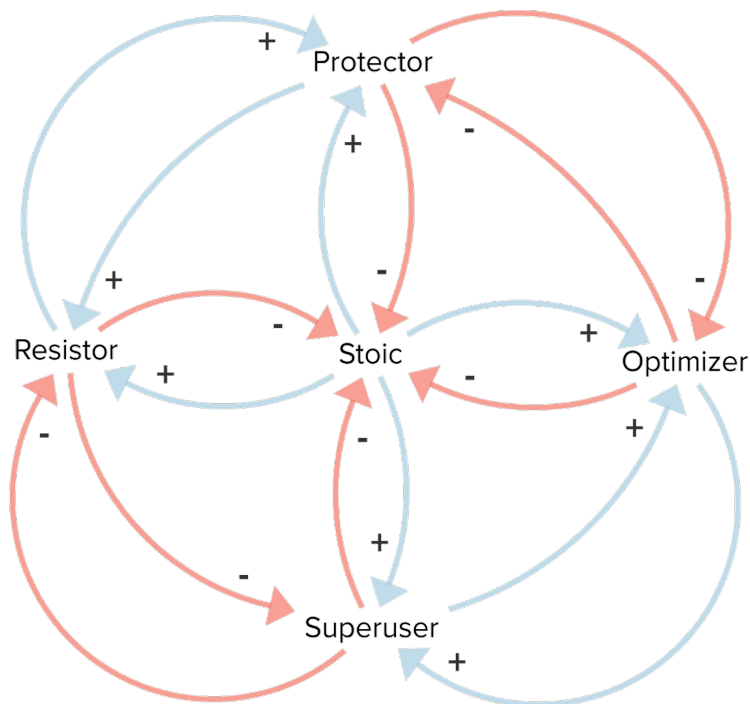

**Fig. 1:** AI-LA Archetypes causal loop diagrams. Blue lines and (+) represent positive polarity and feedback mechanisms. Red lines and (-) represent negative polarity and balancing mechanisms.

In our AI-LA Archetype CLD (see Fig. 1), every archetype has both individual relationships with the others and resides in various multi-archetype feedback loops. Optimizers see AI as a set of tools to enhance efficiency and creativity. They are likely to push for the adoption of AI technologies to optimize work processes, potentially increasing productivity and innovation. Their influence would typically be associated with positive feedback loops in relation with Superusers, accelerating the adoption and development of AI within the landscape architecture profession, and balancing loops in relation with decelerating or neutral forcers like Resistors, Protectors, or Stoics. Resistors view AI as a threat to creative integrity and human value. They may act as a balancing force against the unchecked proliferation of AI technologies by optimizer-led coalitions, advocating for caution and consideration of the broader implications. They contribute to negative feedback loops with Protectors, decelerating the adoption of AI or calling for regulations, and balancing loops with the others.

Stoics are pragmatically indifferent to AI, using it as just another tool without significant enthusiasm or resistance. They represent a stabilizing force in the system, neither accelerating nor decelerating change but maintaining a steady state of use. They create a net balancing loop, providing a counterbalance to both Optimizer and Resistor ends of the spectrum. Superusers are technologically adept and curious, exploring and integrating AI into their practice while sharing their knowledge with others. They are positive feedback for Optimizers, but with a focus on the practical and communal exploration of AI's possibilities, and a balancing force to the others. Protectors are deeply concerned with the ethical, equitable, and environmental implications of AI. They may support AI innovation but are cautious about its impact, ensuring that AI serves a greater societal good. Protectors create negative feedback loops with Resistors (positive polarity) and balancing loops with the others, ensuring that the adoption of AI doesn't compromise ethical standards or societal values.

## 4    Concluding Invitation: Test, Critique, Iterate

We hope the reader pondering the prospect of AI in their firm or classroom might use the AI-LA archetypes model as a resource for switching their mental mode according to context in their research, praxis, or pedagogy. One has the flexibility in this framework to oscillate between mindsets according to the needs of a given situation. An analysis of outdated digital business practices in a firm, for example, might need leaders to put on their Optimizer hats so as to improve workflows using AI-driven tools and free more time for their designers to be focused on what matters most for creative production; then they will need to call on Superuser mode to think about how to actually implement the changes. An update of company policy in reaction to said changes would require the same leaders to take a Protector mindset to ensure equitable labor practices even with these new efficiencies. On the other hand, the same analysis could find a firm "over-digitized" with an overcomplicated tech stack and workflows that bog down designers, in which case taking on the skeptical mentality of the Resistor is useful, finding that indeed their team simply does not want AI to be involved in some aspects of their practice; and so on. Utilizing the AI-LA Archetypes for explorations like these offers a more polymodal view of AI, encouraging landscape architects, scholars, and practitioners to embrace a more fluid, thoughtful approach in integrating AI into their work. We also acknowledge the limitations inherent in this exploration – the rapid evolution of technology, the subjective nature of speculative models, and the complexity of capturing the entire spectrum of a professional field as diverse as landscape architecture. Thus, it is important our model be tested in the wild. We invite readers to utilize AI-LA archetypes in thought experiments, developing technology strategies for their practice, and classroom discussions; to offer critique where the model falls short; and to build on it where they see potential. However it is used, we hope this work is a valuable springboard to more nuanced reflection on landscape architectural practice in an AI-driven world.